\documentclass[a4paper]{article}

\usepackage{INTERSPEECH2020}

\title{
Techniques for Vocabulary Expansion in Hybrid Speech Recognition Systems
}
\name{Nikolay Malkovsky$^{1,2}$, Vladimir Bataev$^{1,4}$, Dmitrii Sviridkin$^1$, Natalia Kizhaeva$^1$, Aleksandr Laptev$^{3}$, Ildar Valiev$^1$ and Oleg Petrov$^1$}
\address{
  $^1$STC-innovations Ltd, St. Petersburg, Russia\\
  $^2$Higher School of Economics, St. Petersburg, Russia\\
  $^3$ITMO University, St. Petersburg, Russia\\
  $^4$University of London, London, UK}
\email{\{malkovskiy, bataev, sviridkin, kizhaeva, laptev, valiev, petrov-o\}@speechpro.com}

\begin{document}

\maketitle

\begin{abstract}

The problem of out of vocabulary words (OOV) is typical for any speech recognition system, hybrid systems are usually constructed to recognize a fixed set of words and rarely can include all the words that will be encountered during exploitation of the system.
One of the popular approach to cover OOVs is to use subword units rather then words. Such system can potentially recognize any previously unseen word if the word can be constructed from present subword units, but also non-existing words can be recognized.
The other popular approach is to modify HMM part of the system so that it can be easily and effectively expanded with custom set of words we want to add to the system.
In this paper we explore different existing methods of this solution on both graph construction and search method levels. We also present a novel vocabulary expansion techniques which solve some common internal subroutine problems regarding recognition graph processing.

\end{abstract}

\noindent\textbf{Index Terms}:
speech recognition, out of vocabulary, weighted finite state transducers, kaldi.

\section{Introduction}

Hybrid speech recognition systems consist of a frame level classifier and a frame sequence combination model. The first part is usually referred to as acoustic model: for acoustic features computed over a fixed period of time it gives the posterior probability over finite set of acoustic units. In the simplest form these acoustic units might be graphemes or phonemes, state of the art systems usually use more advanced tied context-dependent phonemes \cite{young-etal-1994-tree}.
Early on, Gaussian mixture models (GMM) were commonly used as a frame classifiers, but nowadays deep neural networks (DNN) dominate.
The other part of hybrid systems is the hidden Markov model (HMM) that tracks down relations in frame sequences and usually comprises HMM context-dependent phone modeling, word pronunciations and language modeling. Observations for this HMM are per frame acoustic features, transition probabilities in this HMM are frame dependent and defined by acoustic model output for the frame. For the detailed description of a training process refer to \cite{povey2011kaldi}.
Once a system is properly trained the problem of speech recognition can be formulated as finding the best sequence of transitions corresponding to a particular observation sequence. The classical optimal solution to this problem is a dynamic programming based Viterbi algorithm. Although this approach is optimal for HMM paradigm, it is still too expensive for large vocabulary continuous speech recognition (LVCSR) task. This problem lead to development of heuristic search methods. Typically the base search method is a Viterbi beam search which was adopted in token passing method \cite{young1989token} and later developed in weighted finite state transducer (WFST) based decoder \cite{mohri2002weighted} and history conditioned search (HCS) decoder \cite{nolden2017progress}.

One of the problem that arises in such recognition systems is that once fully trained it is still limited to recognize sequences of words over a finite vocabulary. Usually when we are trying to recognize out of vocabulary word the system will try to cover it by a sequence of a shorter words acoustically similar to the origin word, this behaviour may sometimes lead to unwanted results. In some applications it is enough just to detect this occasions \cite{rastrow2009new, white2008confidence, lin2007oov, kumar2012detecting, asami2019recurrent} while other applications require a mechanism to recover out of vocabulary words \cite{thomas2019detection, egorova2018out, kombrink2010recovery, he2014subword, aleksic2015bringing}. For the latter problem there are two major approaches. The first one is usage of subword units rather then words itself. This approach actually improves accuracy overall and partially covers the problem of OOV \cite{thomas2019detection, he2014subword, bisani2005open}.
 \cite{shibata1999byte},
This approach is also the only method of choice for end-to-end speech recognition systems. The other approach is to modify HMM part of the system in such way that the vocabulary can be later extended without acoustic and language models retraining. This approach is mostly researched for WFST based systems in \cite{aleksic2015improved, allauzen2015rapid, bulusheva2016efficient, horndasch2016add, liu2019efficient}. In this paper we present a research on methods described in \cite{aleksic2015improved, liu2019efficient} and present our own improvements of this approach.

The paper is organized as follows: in section $2$ we give basic overview on the usage of WFST in speech recognition; in section $3$ we give an overview of vocabulary expansion approaches based on the WFST framework; in section $4$ we present our own improvements of WFST based vocabulary expansion techniques; in section $5$ we present a summary of our empirical study of the different vocabulary expansion techniques; section $6$ contains a conclusion of the paper and presents research suggestions for our future works.

\section{WFST-based Decoding}

WFSTs is a very effective tool that is commonly used in speech recognition systems to represent different levels of HMM in hybrid systems as a single search network. WFST are easy to work with and have a lot of accumulated knowledge and techniques from finite state machine (FSM) theory and some extensions specifically for weighted transducers. OpenFst open source library \cite{allauzen2007openfst} provides a number of basic utilities for construction and optimization of WFST based search networks.

\subsection{Standard Static Decoding Graph}
Standard decoding graph consists of a composition cascade
$$
H\circ C\circ L\circ G
$$
where
\begin{enumerate}
    \item $H$ transduces context-dependent phoneme HMM model states (or transitions) into context-dependent phonemes
    \item $C$ transduces context dependent phonemes into context independent phonemes
    \item $L$ transduces context independent phonemes into words
    \item $G$ is the acceptor that represents ngram language model
\end{enumerate}
Optimization process may vary for different systems but typically consists of application of the following procedures on intermediate graphs or final $HCLG$ cascade:
\begin{enumerate}
    \item Weight and label pushing
    \item Determinization
    \item Minimization
    \item Local epsilon removal
    \item Connection
\end{enumerate}

Connection removes transitions and states that doesn't belong to some path from initial state to some final state as such paths should not be considered as valid sequences in terminology of WFST/FSM. Minimization and local epsilon removal are used to directly reduce the number of states/arcs in transducer while keeping it equivalent to origin. Determinization is supposed to make transducer deterministic in a sense of FSM (although there are some remarks about epsilon transitions) and can potentially blow up the graph but in most practical cases determinization significantly reduces the number of states in $HCLG$ (and its counterparts containing $L$) and usually helps in the following search process. Weight and label pushing does not change the size of transducer but usually make weights of its paths more smooth which helps in truncated search process.

\subsection{Online Composition with Language Model and Lattice Rescoring}

WFST has a natural way of representing ngram language models which plays important role in hybrid systems as they can drastically improve recognition accuracy. The problem with them is that their research potential hit a dead end two decades ago and the only possibility for improvement of an ngram language model is to add more data which will lead to a larger size of a corresponding transducer.
Complete $HCLG$ graphs are usually several times larger then $G$ and so there is a window due to RAM limitations where we can freely use $G$ but not $HCLG$. There are two major approaches that help in this case. First one is two pass decoding: first pass uses lower order LM to generate intermediate recognition result (lattice) containing best promising hypotheses, second pass applies information from a higher order LM, this second step is usually called ``rescoring'' and can be done with standard WFST composition. The other approach is to use online version of composition. Comparison of these approaches is given in \cite{dixon2012comparison} (although experiments were conducted with GMM acoustic modeling that has some computation specifics compared to DNN). Important issue is that there is a gap between these approaches and baseline $HCLG$ decoding. Surprisingly the ideas from both approaches can be combined and it will be discussed in later sections.

\subsection{Lookahead Composition Pros and Cons}

It was found in \cite{allauzen2009generalized} that online composition needs some additional guidance to perform comparable to offline composition in terms of speed and accuracy. Specifically for composition of $HCL$ and $G$ the following set of so called ``composition filters'' gives comparable with $HCLG$ results:
\begin{enumerate}
    \item Epsilon matching filter
    \item Lookahead reachability filter
    \item Lookahead weight pushing filter
    \item Lookahead label pushing filter
\end{enumerate}

For a detailed description see the original paper. Epsilon matching filter blocks repeated paths following epsilon transitions and leading to the same states. Lookahead reachability filter prevents dead end states (and in some sense takes the role of connection procedure for offline composition) and lookahead weight and label pushing does similar to regular label and weight pushing but using lookahead reachability information.

In some sense HCS decoder can be viewed as WFST decoder that works with separated $HCL$ and $G$. HCS decoder also uses similar lookahead ideas \cite{nolden2011exploiting}.

Unfortunately, usage of lookahead mechanisms described above comes with additional processor time and memory consumption. Usually some kind of caching is used for composition. It is mentioned in \cite{liu2019efficient, allauzen2013pre} that fully expanded cache leads to a huge memory usage (in our experiments 3-5 times larger then $HCLG$ graph) due to internal online composition representation.

Another unavoidable issue arises if we want to isolate search process from composition internals. Essentially state of the composed WFST is a pair of states in the initial WFSTs (one from each transducer). If we want to keep consistency with previously composed states we have to track down the mapping between pair of state to composed state. This for example could be done without saving additional information as follows: if the left WFST has a total of $n$ states, then state ids of graphs to compose being $i$ and $j$ can be uniquely mapped to $jn+i$ ($0$-based indexing assumed). This would usually result in a very sparse indexing while we usually want state indexes to fill interval $[0; M)$ of integers ($M$ being the number of states). The standard method that does this filling is counting how much states we have already composed and map a new composition state to that number (again, $0$-based indexing assumed). This approach requires to keep a mapping that cannot be dropped unless we use external information and thus bounding below the size of cache that needs to be used for composition.

\section{WFST-based Vocabulary Expansion Techniques}

The general mechanism of vocabulary expansion with WFST always involves some modifications of the transducers with symbol table modification. Typically the phoneme set and acoustic modelling stays unchanged and so we only need to modify $L$ and $G$. Modifying $G$ consists of expanding class-based model using $Replace$ operation \cite{aleksic2015improved, bulusheva2016efficient, horndasch2016add, liu2019efficient}. Modifying $L$ might look trivial from the first glance but is actually not a simple problem if we want to keep it minimal \cite{allauzen2015rapid, carrasco2002incremental}. The other problem is tied to composition $CL$. If we construct a $CL'$ for some additional words and try to unite it with origin $CL$ it will result in improper context on the boundaries of new words. In \cite{allauzen2015rapid, bulusheva2016efficient} authors suggested a methods that separated newly added words on groups depending on their boundary phonemes while constructing initial $CLG/HCLG$ with several context-dependent auxiliary symbols that are to be replaced with new words. In \cite{aleksic2015improved} a simpler approach was proposed to deal with context dependency:
\begin{itemize}
    \item Add an auxiliary word for each used phoneme with transcription consisting of this phoneme itself and construct $HCL$ with this modified lexicon. $HCL$ won't be modified later. Prepare transducer $T$ which maps from newly added phoneme words to phonemes
    \item Add a special \$$unknown$ unigram in $G$ to be replaced with new words later
    \item Do whatever routine is needed with $HCL$ and $G$ to be able to effectively compose it later (relabeling, lookahead table construction, sorting \textit{etc})
    \item To add new words construct a lexicon $L'$ over them and use the following recognition graph
    $$
    HCL\circ Replace(G, T\circ L')
    $$
    where $Replace$ replaces occurrences of \$$unknown$ in $G$ with $T\circ L'$
\end{itemize}

This approach was later developed in \cite{liu2019efficient} with some additional WFST composition preinitialization and caching techniques.

\section{Our Contribution}

Our contribution consists of improving OOV handling approach from \cite{aleksic2015improved} with several techniques on the levels of $HCL$ construction, online composition and search algorithms.

\subsection{$HCL$ Construction}

One of the problem with original phoneme-words approach was that lexicon allows words to be separated by optional silence phoneme. In case of word addition it might have a side effect that newly added words might be pronounced with pauses inside the word. To deal with this problem we modify $L$ as follows: typically $L$ has two states (let them be $1$ and $2$) such that word transcriptions are represented as paths from $1$ to $2$ and there are $\langle eps\rangle$ and $sil$ transitions from $2$ to $1$ representing pause/no pause between words. Here rather then adding $1\xrightarrow{}2$ path for phoneme words we add a $1\xrightarrow{}1$ self-loop. On the other hand this approach introduces a problem that no pause is allowed after the newly added words, but this can be easily dealt with later on the level of $G$ if optional silence phone $sil$ is also presented as a phoneme word. It is also notable that if we restrict to use position dependent phonemes then we can use the following routine:
\begin{enumerate}
    \item Add a $1\xrightarrow{}2$ transitions for singleton phonemes
    \item Add a new state $N$
    \item Add transitions $1\xrightarrow{}N$ for each beginning phoneme
    \item Add self-loop $N\xrightarrow{}N$ for each internal phoneme
    \item Add transition $N\xrightarrow{}2$ for each end phoneme
\end{enumerate}
This approach for position dependent phonemes saves some insignificant processing time in the future but doesn't show any improvement of accuracy in our experiments.

\subsection{$T\circ L'$ Construction}

Construction of $T\circ L'$ do not require explicit composition as it is quite trivial to construct the whole composition right away. Note that $L'$ should not cycle itself like lexicon $L$ as we expect it to produce a single word. Like in \cite{liu2019efficient} (but for different reason) we add additional $sil$ transition at the end of the $L'$ representing optional silence after the word to cover the problem described in previous section. In contrast with \cite{liu2019efficient} pre-silence state is also final state so that we can exit $L'$ with or without pause at the end of the word.

\subsection{Kaldi Pseudo-$\langle eps\rangle$ Trick Adaptation for Online Composition}

Although we didn't find any description of it in Kaldi documentation, the following trick is a standard in Kaldi graph construction recipe:
\begin{enumerate}
    \item Relabel input $\langle eps \rangle$ symbols on $G$ (transitions to be relabeled correspond to backoff in LM) with newly introduced symbol $\#0$.
    \item Let's assume that $L$ has the same structure of transcriptions going from $1$ to $2$. Add a self-loop $1\xrightarrow{}1$ with labels $\langle eps \rangle$, $\#0$.
\end{enumerate}
This trick can be viewed as a special implementation of an epsilon matching filter for WFST composition specifically for $L$ and $G$. We found out that this trick also helps standard OpenFst online lookahead composition probably due to the fact that in this case lookahead tables are constructed with $\#0$ being taken into account. We also construct $TL'$ graph and implement $Replace$ operation in a way that follows this epsilon relabeling.

\subsection{Online Composition with $HCLG_k+G_{n-k}$ Transducers}

Inspired by \cite{dixon2012comparison} we were interested whether $HCLG_k+G_{n-k}$ transducers can improve performance in online composition over $HCL+G$ transducers. In particular we explored two boundary cases: $k=1$ and $k=n$. For the first case $HCLG_1$ should be topologically equivalent to $HCL$ and $G_{n-1}$ -- equivalent to $G_n$ and so it is easy to do this ``transition'' of $G_1$ before any other graph preparation routine. In general we found out that due to weight pushing in $HCLG_1$ label and weight pushing lookahead filters can be dropped without accuracy degrading for a medium graph size (up to 200Mb for $HCLG$) but still decreases accuracy for a large graphs (3Gb and more for $HCLG$). Moreover, usage of lookahead pushing filters with $HCLG_1+G_{n-1}$ graphs actually degrades the accuracy. This technique is arguably similar to lookahead tables of lower order LM models used in HCS decoder \cite{nolden2017progress}.

The other case, $k=n$, is interesting due to the fact that it minimizes the right side of the composition. On the left side ($HCL$) we need to somehow push phoneme words through the language model. We achieved it by simply adding phone words as unigrams (although recipe is a little bit more complex if we want to combine it with the technique from the previous section). Note that we assumed that the only nonterminal to be expanded is \$$unknown$ and it is presented only as unigram in $G$. If we had other nonterminals or other occurrences of \$$unknown$ in $G$ we would probably need to add a self-loops for phoneme words at the locations of these transitions but this is out of the scope of the paper.

\begin{table*}[t]
    \caption{WER comparison between different graph composition and preparation routines.}
    \centering
    \begin{tabular}{c|c|c|c|c|c}
        \hline
    \hline
        Dictionary & new words & Graph type & LA reachability & LA weight and label pushing & WER \\
        \hline
         swbd & - & $HCLG$ & - & - & 14.6 \\
         swbd & - & $HCL+G$ & + & + & 14.6 \\
         swbd90 & - & $HCLG$ & - & - & 22.4 \\
         swbd90 & - & $HCL+G$ & + & + & 22.4 \\
         swbd90 & swbd10 & $HCL+G$ & + & + & 19.4 \\
         swbd90 & swbd10 & $HCL+G$ & + & - & 19.6 \\
         swbd90 & swbd10 & $HCL+G$ & - & - & 20.8 \\
         swbd90 & - & $HCLG_1+G_{3-1}$ & + & + & 22.5 \\
         swbd90 & - & $HCLG_1+G_{3-1}$ & + & - & 22.4 \\
         swbd90 & swbd10 & $HCLG_1+G_{3-1}$ & + & - & 19.4 \\
         swbd90 & swbd10 & $HCLG_1+G_{3-1}$ & - & - & 19.4 \\
         swbd90 & - & $HCLG_3+G_0$ & + & - & 22.4 \\
         swbd90 & swbd10 & $HCLG_3+G_0$ & + & + & 20.6 \\
         swbd90 & swbd10 & $HCLG_3+G_0$ & + & - & 19.4 \\
         swbd90 & swbd10 & $HCLG_3+G_0$ & - & - & 19.4 \\
         \hline
    \end{tabular}
    \label{tab:graph_composition_comparison}
\end{table*}

It is also notable that although on the right side of the composition $HCLG_n+G_0$ (with $n=k$) there is a trivial acceptor it is not reducible since it does not contain phone-words which are presented on the left side so we still need it even if we don't add new words. When new words are added $G_0$ plays the role of the container of the lexicon for these new words. One final note concerning $G_0$ graph is that the easiest way to implement it is just straightforwardly construct an acceptor with a single state and loops for every ``real'' word and no weights. This way transducer will have the size of the base vocabulary. Our experiments shows that $HCLG_n+G_0$ without lookahead filters have the same accuracy as $HCL+G$ graphs and thus these types of graphs do not require relabeling that minimizes reachability label representation. This allows us to store the arcs of $G_0$ in a single structure that contains similar arcs with labels in some continuous interval.


\section{Experiments}

All of our experiments are conducted on switchboard benchmark using Kaldi speech recognition toolkit. We use a relatively simple baseline -- TDNN+LSTM acoustic modeling, language model is a 3-gram based on switchboard data without Fisher corpus. WER is reported for Hub'00 evaluation set. For online composition we used a recently added to Kaldi lookahead decoding recipe. Our general experiment approach is as follows: separate words from baseline dictionary into two groups so that the second part is approximately $10\%$ of the whole set, the choice is uniformly random and does not change in all of our experiments. We will refer to the first group as swbd90 and the second group as swbd10. The first group is used to construct a reduced dictionary, the second group is then added via different vocabulary expansion techniques, in all experiments newly added words have a weight of $10$ which might not be optimal but is constant for every presented experiment.

\begin{table}[h]
    \caption{WER Comparison between $L$ modification methods.}
    \centering
    \begin{tabular}{c|c|c|c}
    \hline
    \hline
    dict & new words & method & WER \\
    \hline
    swbd & - & - & $14.6$ \\
    swbd90 & - & - & $22.4$ \\
    swbd90 & - & origin \cite{aleksic2015improved} & $22.4$ \\
    swbd90 & - & L postprocessing & 22.4 \\
    swbd90 & swbd10 & origin \cite{aleksic2015improved} & $19.9$ \\
    swbd90 & swbd10 & L postprocessing & $19.4$ \\
    \hline
    \end{tabular}
    \label{tab:lexicon_comparison}
\end{table}

The first set of experiments compares graph preparation and online composition methods. In particular we experiment with lookahead (LA) usage and LM statistics transferring to the left side of the composition, results are summarized in table \ref{tab:graph_composition_comparison}. In all of these experiments we use methods for $L$ preparation presented in this paper. In general we see that transferring some LM statistics to the left side of composition allows us to turn off some filters without accuracy drop. Also usage of $HCLG_1+G_{3-1}$ with LA weight/label pushing slightly degrades accuracy and this effect was reproduced stably in all of our experiments which our not included in the paper. Another fact that is not represented here is that for a larger graphs accuracy of $HCLG_1+G_{3-1}$ graphs with LA reachability also degraded compared to baseline lookahead while $HCLG_3+G_0$ graphs work stably without LA at all even for a graphs (we conducted experiments with the graphs up to 3Gb).

The second set of experiments compares performance for the system where initial $L$ transducer is constructed from extended lexicon and the system with $L$ transducer being postprocessed as described in the previous section. Note that we use our own implementation of the origin method. These experiments are summarized in table \ref{tab:lexicon_comparison}. In general our experiments show that postprocessing described in our paper has better accuracy then phone word addition. We would also note that Kaldi uses a special silence probability modeling \cite{chen2015pronunciation} which might be messed by our modification schemes but this issue is out of the scope of the paper.

\section{Conclusion and Future Work}

We presented some novel methods on vocabulary expansion in WFST based decoders which reduce the auxiliary expenses for online composition with a detailed comparison between them. We also present some graph preparation routine techniques which improve accuracy when new words are added. We are mostly interested in two improvements in the future works: 1) improve the approach to work with general class based language model expansion; 2) improve recognition of a new words with phone language modeling.

\bibliographystyle{IEEEtran}
\bibliography{bibliography}


\end{document}